\documentclass[cameraready]{Interspeech}
\usepackage{placeins}
\usepackage{fontawesome5}
\usepackage{xcolor}
\usepackage{hyperref}
\usepackage{comment}

\hypersetup{
    colorlinks=true,
    linkcolor=black,
    filecolor=black,
    urlcolor=blue,
}

\title{Huntington Disease Automatic Speech Recognition with Biomarker Supervision}

\author[affiliation={1}]{Charles L.}{Wang$^*$}
\author[affiliation={1}]{Cady}{Chen$^*$}
\author[affiliation={1}]{Ziwei}{Gong}
\author[affiliation={1}]{Julia}{Hirschberg}

\address{
    $^1$ Columbia University \\
    $^*$ \small{Equal contribution}
}

\email{charles.w@columbia.edu, cc4920@columbia.edu, zg2272@columbia.edu, julia@cs.columbia.edu}

\usepackage{comment}

\begin{document}

\maketitle

\begin{abstract}
Automatic speech recognition (ASR) for pathological speech remains underexplored, especially for Huntington’s disease (HD), where irregular timing, unstable phonation, and articulatory distortion challenge current models. We present a systematic HD-ASR study using a high-fidelity clinical speech corpus not previously used for end-to-end ASR training. We compare multiple ASR families under a unified evaluation, analyzing WER as well as substitution, deletion, and insertion patterns. HD speech induces architecture-specific error regimes, with Parakeet-TDT outperforming encoder-decoder and CTC baselines. HD-specific adaptation reduces WER from 6.99\% to 4.95\% and we also propose a method for using biomarker-based auxiliary supervision and analyze how error behavior is reshaped in severity-dependent ways rather than uniformly improving WER. We open-source all code and models.
\end{abstract} 

\begin{flushleft}
\faDatabase\ \textbf{Models:} \href{https://huggingface.co/collections/charleslwang/parakeethd}{charleslwang/parakeethd} \\
\faGithub\ \textbf{Code:} \href{https://github.com/charleslwang/ParakeetHD}{charleslwang/ParakeetHD}
\end{flushleft}

%==================================================================================
% Main paper content (your actual paper)
% Put *all* of your manuscript sections (Intro..Conclusion, incl. figures/tables/equations)
% into content.tex. This keeps main.tex template-compliant and clean.
\section{Introduction}

\subsection{The Gap in Dysarthric ASR}
The clinical landscape of motor-speech disorders is diverse, yet computational research in Automatic Speech Recognition (ASR) has historically operated under the assumption of a dysarthric monolith \cite{Kim2008DysarthricSpeechDatabaseUniversalAccess, Rudzicz2012TORGODatabaseAcousticArticulatoryDysarthria, Qian2023SurveyAutomaticDysarthricSpeechRecognition}. Dominant benchmarks such as UA-Speech and TORGO have successfully driven progress for dysarthric ASR, but these models often fail to generalize to the erratic, hyperkinetic signatures of Huntington’s Disease (HD) \cite{Diehl2019MotorSpeechPatternsHuntington, Rudzicz2012TORGODatabaseAcousticArticulatoryDysarthria, Turrisi2021EasyCallCorpusDysarthricSpeechDataset}. 

HD is characterized by involuntary chorea of the vocal tract, rendering motor-speech patterns in hyperkinetic dysarthria fundamentally distinct from the spastic and hypokinetic patterns typically addressed in the literature \cite{Diehl2019MotorSpeechPatternsHuntington, Geng2020DataAugmentationDisorderedSpeechRecognition}. The degradation of speech in HD is not merely a reduction in signal intensity, but a complex interaction of variable speaking rate, involuntary respiratory bursts, and unpredictable phonatory arrests that differ from more predictable patterns found in other dysarthrias. \cite{Diehl2019MotorSpeechPatternsHuntington, Rudzicz2012TORGODatabaseAcousticArticulatoryDysarthria}. This loss of rhythmic regularity breaks the temporal expectations of modern ASR systems and disrupts token prediction, often resulting in word deletions or alignment failure. \cite{Radford2023RobustSpeechRecognitionWeakSupervision, Diehl2019MotorSpeechPatternsHuntington, Wang2025SelfTrainingWhisperLongDysarthricASR}. 

The lack of specialized high-fidelity corpora has left the HD community under-served, with existing speech-to-text systems remaining largely unaware of the unique bio-acoustic subsystems that drive hyper-kinetic transcription failure. Moreover, prior research in HD-ASR has largely focused on diagnostic classification—using speech to detect the disease—rather than addressing the transcription bottleneck \cite{Subramanian2023DetectingManifestHuntingtonUsingVocalData, Fahed2024LanguageIndependentAcousticBiomarkersHuntington, Takashima2019EndToEndDysarthricSpeechRecognition, Bhat2022ImprovedASRPerformanceDysarthricSpeech}. 
Large-scale end-to-end models such as Whisper provide a robust baseline but lack the nuance required for pathological speech \cite{Radford2023RobustSpeechRecognitionWeakSupervision, Yue2025ProbingWhisperDysarthria}. Recent adaptation efforts, including unified disordered-speech modeling, idiosyncratic atypical-speech modeling, articulatory-aware systems, and parameter-efficient fine-tuning, have sought to bridge this gap \cite{Tobin2025SingleASRDisorderedSpeech, Raja2025IdiosyncraticNormativeAtypicalSpeech, Yue2025AcousticArticulatoryDysarthricSpeechRecognition, Hu2022LoRALowRankAdaptation, Zheng2025SpeechAccessibilityProjectChallenge, Wang2025SelfTrainingWhisperLongDysarthricASR}. 

In this work, we present a systematic study of HD-ASR using a high-fidelity clinical corpus of 94 HD-positive individuals and 36 healthy controls \cite{Subramanian2023DetectingManifestHuntingtonUsingVocalData}. We build on existing research by studying how distinct ASR architectures respond to HD speech and by using clinically grounded biomarker families as auxiliary supervision for adaptation rather than solely as diagnostic endpoints \cite{Radford2023RobustSpeechRecognitionWeakSupervision, Tomanek2021ResidualAdaptersParameterEfficientASR}. Our contributions are as follows:
\begin{enumerate}
    \item \textbf{Cross-Architecture HD-ASR Study:} We compare major ASR families on HD speech and show architecture-specific error regimes.
    \item \textbf{HD-Specific Parameter-Efficient Adaptation:} We adapt Parakeet-TDT to HD speech with encoder-side adapters and evaluate gains in performance and robustness.
    \item \textbf{Biomarker-Informed Auxiliary Supervision:} We test prosodic, phonatory, and articulatory biomarkers as auxiliary supervision for HD-ASR adaptation.
    \item \textbf{Clinical Error Analysis:} We analyze substitutions, deletions, and insertions across model families and severity cohorts.
\end{enumerate}
While we focus on HD, our framework may extend to other atypical speech settings where clinically meaningful acoustic subsystems can be estimated or derived from speech.

\section{HD Speech Corpus and Biomarker Supervision}

This section describes the two ingredients that support our HD-ASR study: the clinical speech corpus used for training and evaluation, and the biomarker-derived feature families used for auxiliary supervision. We first summarize the corpus composition and clinical cohort structure and then describe how we distill clinically motivated prosodic, phonatory, and articulatory measurements into a compact set of supervisory signals for adaptation.

\subsection{Corpus Composition}
We utilize a high-fidelity clinical dataset collected by Beth Israel Deaconess Medical Center (BIDMC) and Canary Speech \cite{Subramanian2023DetectingManifestHuntingtonUsingVocalData, Sierra2026TowardSpeechBasedPremanifestHD}. This corpus consists of 4.5 hours of audio recordings from 130 individuals, of whom 94 are positive for Huntington's Disease (HD) and 36 are healthy controls \cite{Subramanian2023DetectingManifestHuntingtonUsingVocalData}. While our corpus is smaller than widely used dysarthric ASR benchmarks such as TORGO and UA-Speech, to our knowledge it is the first Huntington’s Disease dataset used for end-to-end ASR evaluation and adaptation, and there is currently no publicly available open-source HD speech corpus for ASR. 

The HD cohort is stratified by clinical progression into control, pre-HD, prodromal, and manifest stages based on Unified Huntington's Disease Rating Scale (UHDRS) total motor scores \cite{Subramanian2023DetectingManifestHuntingtonUsingVocalData, Fahed2024LanguageIndependentAcousticBiomarkersHuntington}, from lower to higher severity. Recordings include sustained vowels, syllable repetition (diadochokinetic tasks), prompted responses, and read speech from standard passages such as the Caterpillar passage, providing a comprehensive representation of hyperkinetic motor-speech degradation \cite{Subramanian2023DetectingManifestHuntingtonUsingVocalData, Fahed2024LanguageIndependentAcousticBiomarkersHuntington}.

In addition to the speech corpus itself, our framework requires clinically grounded side information for auxiliary supervision. Prior HD diagnostic systems often rely on expansive feature inventories—frequently exceeding 50 language-independent biomarkers—to capture the broad pathological signature needed for classification \cite{Fahed2024LanguageIndependentAcousticBiomarkersHuntington, Subramanian2023DetectingManifestHuntingtonUsingVocalData}. For ASR adaptation, however, directly importing such a high-dimensional biomarker space can introduce substantial redundancy, multi-collinearity, and reduced interpretability. We therefore construct a compact supervisory representation built from seven interpretable features spanning three core motor-speech subsystems, allowing the auxiliary signal to remain both clinically meaningful and tractable for adaptation.

\subsection{Biomarker Composition}
The selection of these features is directly informed by clinical phonetic literature, which identifies prosodic timing, laryngeal instability, and vowel distortion as primary physiological indicators of HD motor-speech degradation \cite{Diehl2019MotorSpeechPatternsHuntington, Fahed2024LanguageIndependentAcousticBiomarkersHuntington, Romana2020VowelDistortionManifestHuntington}. We utilize \texttt{openSMILE}, \texttt{Parselmouth}, and \texttt{librosa} to extract clinically grounded markers across three subsystems:

\textbf{1. Prosody:}
Following Diehl et al. \cite{Diehl2019MotorSpeechPatternsHuntington}, who established speaking rate and involuntary phonatory arrests as key differentiators of HD motor phenotypes, we extract three temporal and pitch-based features. We utilize a percentile-based dynamic threshold to compute a binary Voice Activity Detection (VAD) indicator, $V(t)$, for each frame $t$, from which we derive (1) a \textit{speech rate proxy} and (2) a \textit{pause-to-speech ratio}:
$$PauseRatio = 1 - \frac{1}{N}\sum_{t=1}^{N} V(t)$$
where $N$ is the total number of frames. To capture macro-fluctuations in vocal fold tension over time, we additionally extract (3) \textit{fundamental frequency variance} ($\sigma(f_0)$) using the \texttt{openSMILE} eGeMAPSv02 feature set.

\textbf{2. Phonation:}
Following Fahed et al. \cite{Fahed2024LanguageIndependentAcousticBiomarkersHuntington}, who demonstrated the diagnostic power of laryngeal micro-perturbation metrics for HD, we use \texttt{Parselmouth} to extract three classical vocal tremor features: (4) \textit{local jitter}, (5) \textit{local shimmer}, and (6) the \textit{Harmonics-to-Noise Ratio (HNR)}. Jitter, representing frequency instability, is computed as the relative mean absolute difference between consecutive fundamental frequency periods:
$$Jitter = \frac{\frac{1}{P-1} \sum_{i=1}^{P-1} |T_i - T_{i+1}|}{\frac{1}{P} \sum_{i=1}^{P} T_i}$$
where $T_i$ represents the duration of the $i$-th extracted pitch period, and $P$ is the total number of periods. Local shimmer is calculated using analogous amplitude differences.

\textbf{3. Articulation:}
Guided by Romana et al. \cite{Romana2020VowelDistortionManifestHuntington} and Perez et al. \cite{Perez2021ArticulatoryCoordination}, who isolate vowel distortion as a premier, early-onset indicator of manifest HD, we design a metric to capture the precision of articulatory targets without requiring manual phonetic alignment. We utilize (7) a \textit{Vowel Space Area (VSA) proxy} derived from the variance of the first two formants ($F_1$ and $F_2$) extracted at 10ms intervals:
$$VSA = \sqrt{\sigma^2(F_1) + \sigma^2(F_2)}$$
This continuous variance metric captures the hypermetric overshoot and target undershoot symptomatic of chorea in the vocal tract.

Continuous measurements across these seven clinically grounded features are z-score normalized against the healthy control cohort and discretized into grouped supervisory labels for downstream biomarker-aware adaptation.

\section{Methodological Framework}

Our framework is organized as a three-stage progression designed to isolate different sources of improvement in HD-ASR. Stage I compares major ASR architecture families under a unified pipeline to identify the strongest zero-shot baseline and characterize architecture-specific failure modes. Stage II adapts that strongest baseline to HD speech using parameter-efficient encoder-side tuning, and Stage III tests whether clinically grounded biomarker supervision provides additional gains or selectively reshapes the resulting error profile. This staged design lets us separate the effect of architecture choice, HD-specific adaptation, and auxiliary clinical supervision.

\subsection{Stage I: Cross-Architecture Evaluation on HD Speech}
Our first objective is to establish how different ASR model families behave on HD speech before adaptation. Rather than assuming that all foundation models fail similarly under pathological speech, we compare encoder-decoder, transducer/TDT, and CTC-based systems under a unified evaluation pipeline. In our study, this architecture suite includes multiple Whisper variants, Parakeet-TDT, and Omnilingual CTC-based ASR. These baselines were chosen to represent widely used open-source ASR families with distinct decoding mechanisms, allowing us to test whether HD speech induces architecture-specific failure modes rather than model-specific errors.
This first stage is designed to answer two questions: (1) which architectures are most robust in zero-shot HD-ASR, and (2) whether different models fail through distinct error regimes. To that end, we evaluate all systems on the identical held-out test set using the same transcript normalization and WER computation pipeline, and decompose WER into substitutions, deletions, and insertions. This enables us to move beyond raw aggregate error and identify whether models preserve lexical coverage, hallucinate content, or collapse through omission under hyperkinetic disruption.

\subsection{Stage II: HD-Specific Parameter-Efficient Adaptation of Parakeet}
Having identified Parakeet-TDT 0.6B as the strongest zero-shot baseline in Stage I, the second stage adapts it to the HD corpus itself. Rather than updating all model weights, we perform parameter-efficient adaptation of Parakeet-TDT using encoder-side adapters trained on the HD training split of the corpus.
This experiment serves two purposes. First, it establishes whether HD-specific adaptation can improve upon an already strong zero-shot recognizer. Second, it allows us to analyze how adaptation changes the model’s error profile across clinical severity groups. We therefore compare zero-shot Parakeet and HD-adapted Parakeet not only by overall WER, but also by cohort-level performance and error composition.

\subsection{Stage III: Biomarker-Informed Auxiliary Supervision}
The final stage builds on the HD-adapted Parakeet model from Stage II and asks whether clinically grounded acoustic biomarkers can further improve adaptation when used as auxiliary supervision.
Rather than treating biomarkers as decoder-side textual prefixes, we use them as encoder-level supervisory signals attached to the same utterances used for ASR training.
For each utterance, continuous biomarkers are z-normalized against training controls, discretized into low/medium/high-style bins, combined into a family-level label, and predicted from masked mean-pooled encoder states via a linear head trained jointly with the ASR objective.
Concretely, we construct three auxiliary supervision settings corresponding to the major motor-speech subsystems in HD:
\begin{itemize}
    \item \textbf{Prosody}: speech rate proxy, pause ratio, and pitch variance
    \item \textbf{Phonation}: jitter, shimmer, and HNR
    \item \textbf{Articulation}: VSA proxy
\end{itemize}

\begin{figure}[t]
    \centering
    \includegraphics[width=0.70\columnwidth]{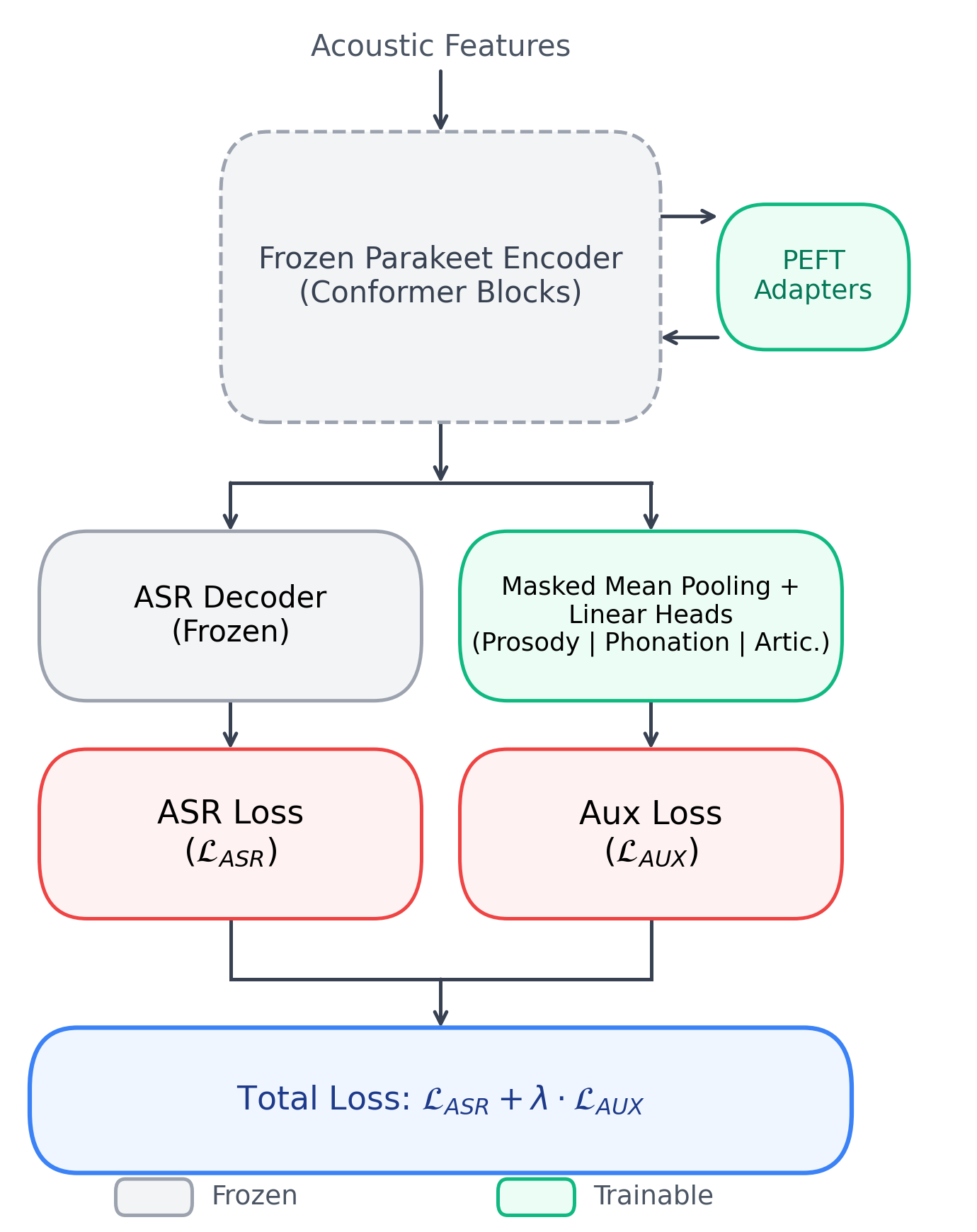}
    \caption{Overview of our adapter-based HD-ASR framework. A frozen Parakeet-TDT encoder is augmented with trainable PEFT adapters. Standard ASR supervision is retained, while a masked mean-pooled encoder representation is used to predict one auxiliary biomarker family per run (prosody, phonation, or articulation).}
    \label{fig:parakeet_aux}
\end{figure}
For each setting, the adapter-augmented Parakeet model is trained on the standard transcription objective while simultaneously predicting the corresponding biomarker-group labels from masked mean-pooled encoder representations. In this way, the main task remains ASR, but the encoder is encouraged to organize its internal representation around clinically meaningful motor-speech structure.
This design allows us to test whether biomarkers act as useful side information for HD-ASR, whether different biomarker families contribute differently to adaptation, and whether clinically grounded auxiliary supervision changes overall performance or selectively reshapes where the model improves.

\subsection{Experiments Setup and Data Partitioning}
To ensure rigorous evaluation and prevent data leakage, the corpus is partitioned into speaker-independent training, validation, and test sets using a \textbf{70/10/20} split. Partitioning is strictly stratified by clinical cohort—Healthy Control, Pre-manifest, Prodromal, and Manifest HD—to maintain a balanced motor-severity distribution across all folds and ensure robust generalization to hyperkinetic speech.

\textbf{Baseline Evaluation:} All zero-shot models are evaluated on the same held-out test split using a unified transcript normalization pass. For long-form capable models, inference is performed using each model’s appropriate decoding strategy while maintaining a common scoring pipeline.

\textbf{Parakeet Adapter-Based Adaptation:} We adapt Parakeet on the HD corpus using parameter-efficient encoder-side adapters under standard ASR supervision, while keeping the pretrained backbone frozen. Models were adapted from NVIDIA Parakeet-TDT 0.6B v2 using AdamW with a learning rate of $3 \times 10^{-4}$, weight decay of $1 \times 10^{-3}$, mixed-precision training, gradient accumulation of 4 steps, duration-based Lhotse batching with a training batch duration of 20 seconds, dynamic bucketing, and subsampling-convolution chunking for long-form audio stability. Training was conducted on $2\times$ H200 GPUs for 8 epochs on Parakeet-HD and 5 epochs for the auxiliary models.

Because a framework-level validation issue in the current Parakeet implementation prevented stable validation-based checkpoint selection, all adapter-based experiments were run with fixed training schedules and final checkpoints were used for downstream evaluation.

\textbf{Biomarker-Auxiliary Adapter-Based Adaptation:} For biomarker-aware variants, we initialize from the HD-adapted Parakeet adapter checkpoint and continue training with an auxiliary classification loss corresponding to one of the three biomarker families. Each auxiliary run uses the same optimizer, learning rate, weight decay, device count, gradient accumulation, duration-based batching, and subsampling-convolution chunking settings as the plain HD adaptation stage, with $\lambda = 0.1$ for the auxiliary loss and a fixed training schedule of 10 epochs. The total loss takes the form:
\begin{equation}
\mathcal{L}_{total} = \mathcal{L}_{ASR} + \lambda \mathcal{L}_{bio}
\end{equation}
where $\mathcal{L}_{bio}$ denotes the auxiliary supervision term for one biomarker family and $\lambda$ controls the tradeoff between transcription fidelity and biomarker-aware representation learning.
\section{Results and Discussion}

This section reports the main empirical findings of our three-stage framework and interprets what they reveal about HD-ASR. We first compare zero-shot ASR architectures, then evaluate HD-specific adaptation and biomarker-aware supervision, and finally analyze how these interventions reshape error behavior across clinical severity.

\FloatBarrier
\subsection{Cross-Architecture Baselines}
We first compare zero-shot ASR model families on the HD test set. Table~\ref{tab:baseline_architectures}
reports overall WER across model families, while Table~\ref{tab:baseline_error_composition}
summarizes how each baseline distributes its total errors across substitutions, deletions, and insertions.

\begin{table}[!t]
\centering
\caption{Zero-shot ASR performance across model families on the HD test set. Results are reported on 47 scored test utterances.}
\label{tab:baseline_architectures}
\resizebox{\columnwidth}{!}{%
\begin{tabular}{lccc}
\hline
\textbf{Model} & \textbf{Architecture Family} & \textbf{Parameters} & \textbf{WER (\%)} \\ \hline
Whisper-small & Encoder-decoder & 244M & 19.78 \\
Whisper-medium & Encoder-decoder & 769M & 26.91 \\
Whisper-large-v2 & Encoder-decoder & 1.55B & 18.44 \\
Parakeet-TDT 0.6B & TDT / Transducer & 600M & \textbf{6.99} \\
Meta Omnilingual 300M & CTC & 300M & 30.46 \\ \hline
\end{tabular}%
}
\end{table}

\begin{table}[!t]
\centering
\caption{Error composition across zero-shot ASR baselines on HD speech. Values are reported as percentages of each model's total errors $(S + D + I)$.}
\label{tab:baseline_error_composition}
\resizebox{\columnwidth}{!}{%
\begin{tabular}{lccc}
\hline
\textbf{Model} & \textbf{Substitution Share (\%)} & \textbf{Deletion Share (\%)} & \textbf{Insertion Share (\%)} \\ \hline
Whisper-small & 17.99 & 9.35 & \textbf{72.66} \\
Whisper-medium & 13.29 & 6.68 & \textbf{80.04} \\
Whisper-large-v2 & 16.56 & 9.18 & \textbf{74.27} \\
Parakeet-TDT 0.6B & \textbf{41.90} & 29.68 & 28.43 \\
Meta Omnilingual 300M & \textbf{43.30} & 22.85 & 33.85 \\ \hline
\end{tabular}%
}
\end{table}

Table~\ref{tab:baseline_architectures} shows a strong architectural gap on HD speech: Parakeet-TDT reaches 6.99\% WER, far ahead of Whisper-large-v2 at 18.44\% and the CTC baseline at 30.46\%. The Whisper family is also non-monotonic, with Whisper-medium underperforming Whisper-small, suggesting that robustness to hyperkinetic speech is not explained by model scale alone. Table~\ref{tab:baseline_error_composition} further shows that the baselines fail differently: Whisper is strongly insertion-dominated (72.66--80.04\% of total errors), whereas Parakeet is markedly more balanced. This means that the same HD utterances are not simply becoming uniformly harder for all systems; rather, different architectures break down in different ways, with encoder-decoder models tending to overgenerate while Parakeet better preserves coverage. HD speech therefore exposes architecture-specific error regimes rather than a uniform increase in difficulty.

\FloatBarrier
\subsection{Biomarker-Informed Auxiliary Supervision}
We next evaluate whether HD-specific adaptation and biomarker-aware auxiliary supervision improve
recognition performance. Table~\ref{tab:biomarker_aux_main} summarizes overall results.
\begin{table}[!t]
\centering
\caption{ASR performance for biomarker-auxiliary Parakeet variants.}
\label{tab:biomarker_aux_main}
\resizebox{\columnwidth}{!}{%
\begin{tabular}{lcccc}
\hline
\textbf{Model} & \textbf{WER (\%)} & \textbf{Substitutions (\%)} & \textbf{Deletions (\%)} & \textbf{Insertions (\%)} \\ \hline
Parakeet-HD & \textbf{4.95} & 2.09 & \textbf{1.29} & 1.57 \\
Parakeet-HD-Prosody & 6.11 & 2.13 & 2.53 & 1.45 \\
Parakeet-HD-Phonation & 6.07 & \textbf{1.92} & 2.77 & 1.38 \\
Parakeet-HD-Articulation & 6.44 & 1.94 & 3.21 & \textbf{1.29} \\ \hline
\end{tabular}%
}
\end{table}
Table~\ref{tab:biomarker_aux_main} shows that plain HD-specific adaptation is best overall, reducing WER from 6.99\% to 4.95\% while improving substitutions, deletions, and insertions simultaneously. None of the biomarker-aware variants exceeds this result. However, the auxiliaries do alter the error profile in a structured way: phonation yields the lowest substitution rate, articulation the lowest insertion rate, and all three incur a deletion penalty. The main effect of biomarker supervision is therefore not uniform WER improvement, but a shift toward more conservative decoding.

\FloatBarrier
\subsection{Error-Type and Severity Analysis}
To understand how auxiliary supervision changes recognition behavior, we compare each biomarker-aware
variant against Parakeet-HD using cohort-level percentage-point deltas in WER and error types.
\begin{table}[!t]
\centering
\caption{Cohort-level change relative to Parakeet-HD for biomarker-auxiliary variants. Values are percentage-point differences; negative is better. Best $\Delta$WER in each cohort is bolded.}
\label{tab:final_delta_by_severity}
\resizebox{\columnwidth}{!}{%
\begin{tabular}{lccccc}
\hline
\textbf{Model} & \textbf{Stage} & $\Delta$ \textbf{WER} & $\Delta$ \textbf{Sub.} & $\Delta$ \textbf{Del.} & $\Delta$ \textbf{Ins.} \\ \hline
Parakeet-HD-Prosody & Control & \textbf{-0.26} & -0.05 & +0.16 & -0.37 \\
Parakeet-HD-Prosody & Pre-HD & \textbf{-0.22} & -0.11 & +0.00 & -0.11 \\
Parakeet-HD-Prosody & Prodromal & -0.24 & +0.00 & -0.24 & +0.00 \\
Parakeet-HD-Prosody & Manifest & +3.59 & +0.19 & +3.35 & +0.05 \\ \hline

Parakeet-HD-Phonation & Control & -0.05 & -0.16 & +0.37 & -0.26 \\
Parakeet-HD-Phonation & Pre-HD & -0.11 & +0.00 & +0.00 & -0.11 \\
Parakeet-HD-Phonation & Prodromal & \textbf{-0.37} & +0.00 & -0.24 & -0.12 \\
Parakeet-HD-Phonation & Manifest & +3.30 & -0.33 & +3.83 & -0.19 \\ \hline

Parakeet-HD-Articulation & Control & -0.16 & +0.00 & +0.10 & -0.26 \\
Parakeet-HD-Articulation & Pre-HD & \textbf{-0.22} & -0.11 & +0.00 & -0.11 \\
Parakeet-HD-Articulation & Prodromal & +3.18 & -0.49 & +4.77 & -1.10 \\
Parakeet-HD-Articulation & Manifest & \textbf{+3.06} & -0.19 & +3.30 & -0.05 \\ \hline
\end{tabular}%
}
\end{table}
Table~\ref{tab:final_delta_by_severity} shows that biomarker supervision can slightly help in milder cohorts: prosody is best on control, prosody and articulation tie on pre-HD, and phonation is best on prodromal speech. In manifest HD, however, all auxiliary variants degrade substantially relative to Parakeet-HD, with WER increases of +3.06 to +3.59 points. These losses are driven mainly by deletions rather than insertions, and the same pattern appears for articulation in prodromal speech. This pattern suggests that auxiliary biomarker objectives encourage the encoder to prioritize clinically meaningful structure over lexical coverage. While this improves precision in milder speech, it backfires in severe HD by forcing the model into overly conservative decoding—resulting in word omissions rather than hallucinations. Overall, biomarker supervision appears most helpful when pathology remains structured enough to support lexical recovery, but counterproductive when severe disruption makes conservative decoding collapse into omission.

These results should be interpreted in light of several limitations. The corpus is high quality but relatively small, consists primarily of controlled and read speech, and does not fully cover the most severe end of the HD spectrum. In addition, our biomarker supervision is intentionally lightweight, relying on discretized auxiliary labels and a fixed loss weighting, and our adapter-based runs used fixed training schedules because stable validation-based checkpoint selection was unavailable in the current Parakeet framework. Future work should test broader and more severely impaired HD cohorts, extend evaluation to spontaneous and conversational speech, and explore richer biomarker fusion strategies aimed at reducing omission-driven failure under advanced hyperkinetic disruption.

\FloatBarrier

\section{Conclusion}
In this work, we presented a systematic study of automatic speech recognition for Huntington’s Disease using a high-fidelity clinical corpus and a unified comparison framework. We showed that HD speech exposes architecture-specific failure modes rather than simply increasing overall difficulty: Parakeet-TDT was substantially more robust than the encoder-decoder and CTC baselines, and its errors were markedly less insertion-dominated. We further showed that HD-specific parameter-efficient adaptation provides the strongest overall performance, reducing WER from 6.99\% to 4.95\% while improving substitutions, deletions, and insertions simultaneously. Finally, we found that biomarker-informed auxiliary supervision does not uniformly improve ASR, but instead reshapes the error profile in a clinically meaningful way.

%==================================================================================
% IMPORTANT LENGTH/PAGE RULES FROM TEMPLATE:
% - Regular papers: max 6 pages total; pages 5-6 reserved exclusively for:
%   Acknowledgments, Generative AI Use Disclosure, References (may begin earlier if space permits).
%   No other content may appear on pages 5-6.
%   Any appendices must be within the first four pages.
% - Long papers: max 10 pages total; pages 9-10 reserved exclusively for:
%   Acknowledgments, Generative AI Use Disclosure, References (may begin earlier if space permits).
%   No other content may appear on pages 9-10.
%   Any appendices must be within the first eight pages.
%
% Therefore:
% - If you include an appendix, it MUST be placed such that it stays within the allowed
%   first pages (<=4 for regular, <=8 for long). This is an author responsibility: LaTeX
%   cannot enforce this without risky template modifications.
%
% The appendix is placed here (before Acknowledgments/AI disclosure/References) so it can
% still fall within the allowed first pages when you manage length appropriately.

%==================================================================================
% Acknowledgments (camera-ready only; do not include in the review submission PDF)

\section{Acknowledgments}

We would like to thank Beth Israel Deaconess Medical Center (BIDMC) and Canary Speech for providing the clinical speech corpora used in this study. CW was supported by a grant from the Columbia Data Science Institute. ZG was supported by the National Science Foundation and by DoD OUSD (R\&E) under Cooperative Agreement PHY-2229929 (The NSF AI Institute for Artificial and Natural Intelligence).

%==================================================================================
% References

\bibliographystyle{IEEEtran}
\bibliography{refs}

\end{document}